\documentclass[sigconf]{acmart}
\usepackage{amssymb}
\usepackage{pifont}
\usepackage{multirow}
\usepackage{booktabs}       
\usepackage{multirow}       
\usepackage{graphicx}      
\usepackage{xcolor}        
\usepackage{colortbl}    
\usepackage{array}         
\usepackage{enumitem}
\usepackage{amsmath}
\usepackage{dsfont}
\usepackage{hyperref}
\usepackage{subcaption}
\usepackage{balance}

\AtBeginDocument{%
  }

\copyrightyear{2026}
\acmYear{2026}
\setcopyright{cc}
\setcctype{by-nc-nd}
\acmConference[WWW '26] {Proceedings of the ACM Web Conference 2026}{April 13--17, 2026}{Dubai, United Arab Emirates.}
\acmBooktitle{Proceedings of the ACM Web Conference 2026 (WWW '26), April 13--17, 2026, Dubai, United Arab Emirates}
\acmISBN{979-8-4007-2307-0/2026/04}
\acmDOI{10.1145/3774904.3792336}

\begin{document}

\title{LoSemB: Logic-Guided Semantic Bridging for Inductive Tool Retrieval}

\author{Luyao Zhuang}
\email{luyao.zhuang@connect.polyu.hk}
\orcid{0009-0007-1653-9843}
\affiliation{%
  \institution{The Hong Kong Polytechnic University}
  \city{Hong Kong SAR}
  \country{China}
}

\author{Qinggang Zhang}
\authornote{Corresponding author}
\email{qinggangg.zhang@connect.polyu.hk}
\orcid{0000-0002-0247-4942}
\affiliation{%
  \institution{The Hong Kong Polytechnic University}
  \city{Hong Kong SAR}
  \country{China}
}

\author{Huachi Zhou}
\email{huachi.zhou@connect.polyu.hk}
\orcid{0000-0002-8301-8470}
\affiliation{%
  \institution{The Hong Kong Polytechnic University}
  \city{Hong Kong SAR}
  \country{China}
}

\author{Yujing Zhang}
\email{yu-jing.zhang@connect.polyu.hk}
\orcid{0009-0004-2466-3164}
\affiliation{%
  \institution{The Hong Kong Polytechnic University}
  \city{Hong Kong SAR}
  \country{China}
}

\author{Xiao Huang}

\email{xiao.huang@polyu.edu.hk}
\orcid{xxxx-xxxx-xxxx-xxxx}
\affiliation{%
  \institution{The Hong Kong Polytechnic University}
  \city{Hong Kong SAR}
  \country{China}
}

\renewcommand{\shortauthors}{Luyao Zhuang, Qinggang Zhang, Huachi Zhou, Yujing Zhang, \& Xiao Huang}

\begin{abstract}
Equipping large language models (LLMs) with external tools has emerged as a promising paradigm for addressing real-world tasks. Nonetheless, with the web-based tool ecosystems rapidly expanding, it is impractical to include all tools within the limited input length of LLMs. To alleviate these issues, researchers have explored incorporating a tool retrieval module to select the most relevant tools or represent tools as unique tokens within LLM parameters. However, most state-of-the-art methods are under transductive settings, assuming all tools have been observed during training. Such a setting deviates from reality as tools on the web are constantly updated and new tools are frequently added to the online ecosystem. When dealing with these unseen tools, which refer to tools not encountered during the training phase, these methods are limited by two key issues, including the large distribution shift and the sensitivity of semantic-only retrieval. To this end, inspired by human cognitive processes of mastering unseen tools through discovering and applying the logical information from prior experience, we introduce a novel \underline{\textbf{Lo}}gic-Guided \underline{\textbf{Sem}}antic \underline{\textbf{B}}ridging framework for inductive tool retrieval, namely, \textbf{\texttt{LoSemB}}, which aims to mine and transfer latent logical information for inductive tool retrieval without costly retraining. Specifically, \texttt{LoSemB} contains a logic-based embedding alignment module to mitigate distribution shifts and a relational augmented retrieval mechanism to overcome the limitations of semantic-only similarity methods. Extensive experiments demonstrate that \texttt{LoSemB} achieves advanced performance in both the inductive and transductive settings. 
% while maintaining desirable effectiveness in the transductive setting.
\end{abstract}




\maketitle

\section{Introduction}

\begin{figure}[t]
  \centering
  \includegraphics[width=0.96\linewidth]{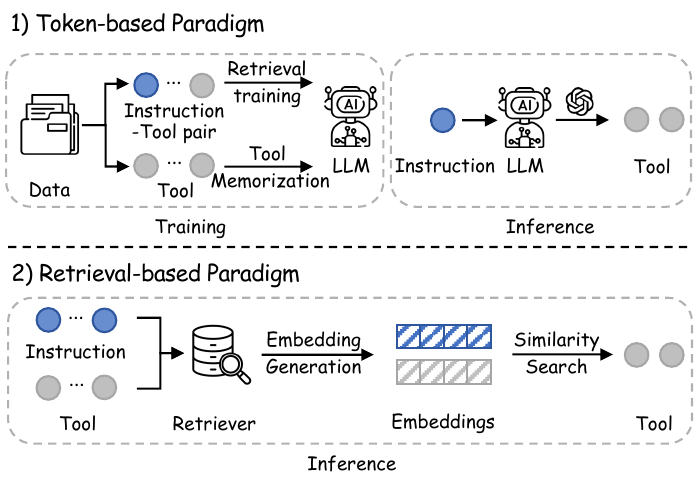}
  \vspace{-4mm}
\caption{Token-based vs. Retrieval-based paradigm.}
\vspace{-6mm}
\label{fig:orig_method}
\end{figure}

While large language models (LLMs)~\cite{achiam2023gpt,anthropic2024claude,liu2024deepseek,touvron2023llama} have demonstrated remarkable capabilities across diverse tasks~\cite{kung2023performance,zha2025data}, they still fall short in some problems, \textit{e.g}, complex computations and providing real-time information~\cite{srivastava2022beyond}, due to their reliance on fixed and parametric knowledge~\cite{qu2025tool}. Recently, to extend the abilities of LLMs, tool learning~\cite{cai2024large,chen2024smurfs,lu2023chameleon,lu2024chameleon,qu2024towards}, which augments LLMs with external tools, has attracted enormous attention. For instance, web search engines enable LLMs to access real-time information, overcoming temporal limitations and providing informative responses.

\begin{figure*}[t]
    \centering
    \begin{subfigure}[b]{0.25\textwidth}
        \centering
        \includegraphics[width=\textwidth]{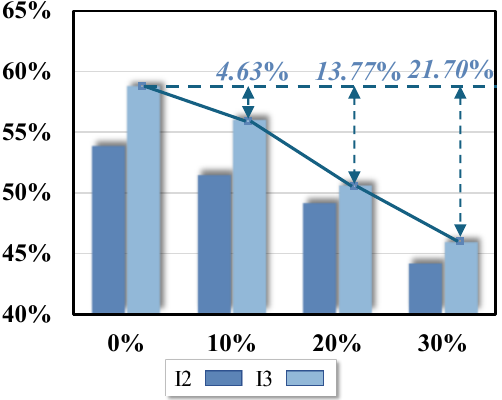}
        \caption{Impact of Unseen Tool Ratio.}
        \label{fig:unseen_ratio}
    \end{subfigure}
    \hfill
    \begin{subfigure}[b]{0.23\textwidth}
        \centering
        \includegraphics[width=0.88\textwidth]{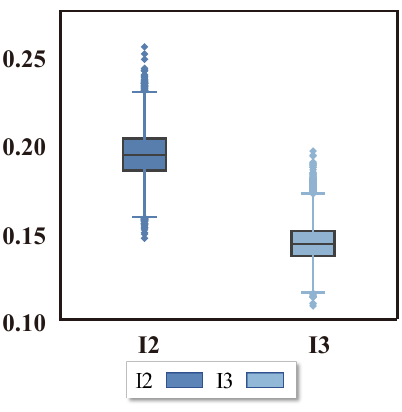}
        \caption{KL Divergence Distributions.}
        \label{fig:kl_divergence}
    \end{subfigure}
    \hfill
    \begin{subfigure}[b]{0.25\textwidth}
        \centering
        \includegraphics[width=\textwidth]{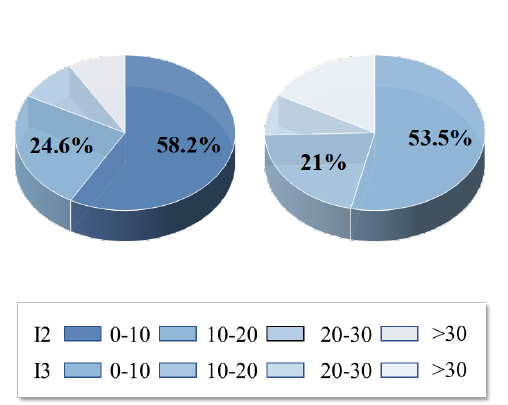}
        \caption{Tool Co-occurrence Analysis.}
        \label{fig:co_occurrence}
    \end{subfigure}
    \hfill
    \begin{subfigure}[b]{0.25\textwidth}
        \centering
        \includegraphics[width=\textwidth]{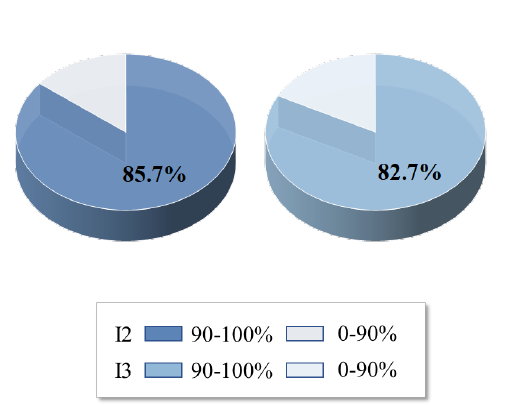}
        \caption{Tool Overlap Analysis.}
        \label{fig:tool_overlap}
    \end{subfigure}
    \caption{\textbf{(a)} Impact of unseen tool ratio on retrieval performance, \textbf{(b)} Comparison of KL divergence distributions between seen and unseen tools, \textbf{(c)} Distribution of tool co-occurrence counts across datasets, \textbf{(d)} Tool overlap percentages between each instruction's tool set and its Top-5 semantically similar instructions. All results are based on ToolBench (I2) and (I3).}
    \label{fig:pre_exp}
    % \vspace{-4mm}
\end{figure*}

% \begin{figure*}[t]
%     \centering   
%     \includegraphics[width=1\textwidth]{figure/pre_exp.pdf}    
   
%      \caption{\textbf{(a) Impact of unseen tool ratio on retrieval performance}, \textbf{(b) Comparison of KL divergence distributions} calculating the difference between training and unseen distributions for instructions and tools, \textit{i.e.}, "seen" refers to ID (in-distribution) data while "unseen" corresponds to OOD (out-of-distribution) data, \textbf{(c) Analysis of tool co-occurrence number} that illustrates the distribution of tool co-occurrence counts across datasets, \textbf{(d) Analysis of tool overlap in semantically similar Instructions} that reveals the distribution of overlap percentages between each instruction's tool set and the combined tool set derived from its top5 most semantically similar instructions.}
%     \vspace{-0.45cm}
%      \label{fig:pre_exp}
% \end{figure*}   

However, as the number of tools equipped with LLMs increases to the tens of thousands, it has become challenging to contain all the tools and their descriptions~\cite{hsieh2023tool}, hindered by the limited context length of LLMs~\cite{brown2020language,liu2024lost}. To tackle this issue, two primary research directions have emerged, as depicted in Figure \ref{fig:orig_method}: 1) Token-based methods~\cite{hao2023toolkengpt,schick2023toolformer,wang2024toolgen} represent each tool as a specific token and integrate tool knowledge directly into the LLM's parameters. Through fine-tuning processes that memorize each tool document and map it to its corresponding token, these models develop the capability to generate appropriate tool calls autonomously. 2) Retrieval-based methods~\cite{qu2024towards,xu2024enhancing,zhang2024data} retrieve relevant tools from the large tool repository by calculating the similarity between the instruction and the tools. These methods employ either sparse lexical similarity retrievers like BM25~\cite{robertson2009probabilistic}, which require no training, or dense embedding retrievers that utilize pre-trained language models (PLMs). Dense retrievers~\cite{zhao2024dense} can be further categorized into training-free approaches, such as Re-Invoke~\cite{chen2024re}, which rewrites instructions and extracts their intents to enhance tool documentation, and training-based approaches like ToolRetriever~\cite{qintoolllm}, which fine-tunes on instruction-tool pairs to optimize retrieval performance.

In this regard, most state-of-the-art methods require domain-specific training and operate under transductive settings, where all tools are available during the training phase. However, real-world scenarios frequently involve new tools or the addition of new functionalities to existing tools, posing challenges for both paradigms. For token-based methods, the core mechanism relies on learning compressed token representations that associate each tool with its usage contexts through training. When encountering unseen tools, these methods become completely ineffective unless the entire model is retrained to learn new token associations. On the other hand, retrieval-based methods struggle to generate accurate representations for unseen tools, as evidenced by our preliminary experiments showing large performance degradation for fine-tuned retrievers in the inductive setting.

Based on the experimental results, we identified two challenges in retrieval-based methods when handling unseen tools. \textbf{\ding{182} Large Distribution Shift.} Our analysis reveals that unseen tools show large distribution shifts due to functional diversity across different tools. This causes retrievers trained on seen tools to learn biased representations that fail to capture the true functionality of unseen tools, ultimately resulting in retrieval failures. \textbf{\ding{183} Sensitivity of Semantic-only Retrieval.} Traditional retrieval-based methods primarily rely on computing semantic similarity between instructions and tool descriptions. However, this approach is highly sensitive to the quality of learned representations. When dealing with unseen tools, their representations suffer from distribution shifts, leading to worse retrieval performance. Drawing from practical experience, we propose and validate that tools typically exhibit sparse co-occurrence relationships, and semantically similar instructions often correspond to overlapping tool sets. This logical information provides opportunities to mitigate distribution shifts and achieve more robust retrieval.

To this end, we draw inspiration from how humans adapt to unseen tools. When encountering unseen tools, humans first systematically organize their existing knowledge, identifying the relationships between tools and their usage scenarios, as well as the functional connections among tools. Through this structured analysis, humans extract deeper logical information and then transfer it to guide their understanding and use of the unseen tool~\cite{goldstone2012introduction,winston1980learning}. This cognitive process reveals that logical information plays a crucial role in adapting to unseen tools. Thus, our framework is guided by two critical questions: \ding{182} How can we efficiently extract hidden logical information and transfer it to learn better representations for unseen tools? \ding{183} How can we effectively integrate logical information into the retrieval paradigm, rather than relying only on semantic similarity, to guide an accurate tool retrieval process? Our main contributions are summarized as follows: 

% Based on our preliminary findings, logical information serves as a crucial component for addressing the challenges in inductive tool retrieval. Accordingly, our framework centers on two critical questions: \ding{182} How can we extract hidden logical information and transferring them to learn better representations for unseen tools? \ding{183} How can we effectively integrate logical information into the retrieval paradigm, rather than relying only on semantic similarity, to guide an accurate tool retrieval process?

\begin{itemize}[leftmargin=*] 
    \item We identify the key challenges of existing tool retrieval methods in the inductive setting and propose \texttt{LoSemB} to improve the accuracy of retrieving unseen tools without costly retraining.
    \item \texttt{LoSemB} introduces a novel logic-based embedding alignment module that addresses the distribution shift by integrating logical features into the representations of unseen tools.
    \item Building upon logically enhanced embeddings, \texttt{LoSemB} further adopts a relational augmented retrieval mechanism that leverages both logical pruning and similarity of the enhanced-embeddings to overcome the sensitivity of semantic-only retrieval.
    \item Experiments show that \texttt{LoSemB} achieves advanced performance in inductive settings while maintaining desirable effectiveness in the transductive setting.
\end{itemize}

\section{Problem Statement}

Given a test instruction $q_{test}$, the tool retrieval task requires identifying a tool set $\mathcal{T}_{\mathrm{ret}} \subseteq \mathcal{T} = \{t_j\}_{j=1}^{m}$ with the highest relevance scores calculated by a retrieval function $f(\cdot)$. In this paper, we delve into the impact of both transductive and inductive settings on tool retrieval. The transductive setting assumes all tools in the repository are seen during training, where we use instruction-tool pairs to finetune the retriever. However, in practice, tool repositories are frequently updated with new tools that were unavailable during training. We define this scenario as the inductive tool retrieval setting. Therefore, when training the retriever for an inductive setting, we must filter the training dataset to exclude the unseen tools and their corresponding unseen instructions to properly evaluate tool retrieval performance in the inductive setting.

% The retrieval function $f(\cdot)$ first transforms $q^{\mathrm{test}}$ and each $t_j$ into embeddings, then computes similarity scores between them. 

\section{Preliminary Study}

Before going into the technical details of \texttt{LoSemB}, we first conduct a preliminary study to identify the primary challenges of the retrieval-based method in the inductive setting.

\subsection{Performance Degradation}

While retrieval-based methods show promise for tool learning, their effectiveness in handling unseen tools remains unclear. To investigate this, we evaluate fine-tuned retrievers in the inductive setting. As shown in Figure \ref{fig:pre_exp} (a), we trained a BERT-base~\cite{devlin2019bert} retriever on the I2 and I3 training sets of ToolBench and tested it with varying proportions of unseen tools (10\%, 20\%, and 30\%). Our experiments reveal that retrieval performance declines as the proportion of unseen tools increases. Notably, the I3 dataset showed significant degradation, with relative accuracy drops of 4.63\%, 13.77\%, and 21.70\% for the three unseen tool ratios, respectively.

\subsection{Cause Analysis}
\label{sec:kl}

To understand what underlying challenges lead to this performance degradation, we conduct analyses about the representations of unseen tools and the impact of the existing retrieval architecture, as these two critical factors directly determine retrieval performance.

\textbf{Challenge \ding{182}: Large Distribution Shift.} As shown in Figure \ref{fig:pre_exp} (b), we measured the KL divergence between the representation distributions of unseen tools \textit{v.s.} seen tools (tools encountered during training). We observe that unseen tools exhibit large distributional divergence from seen tools. To understand the impact of this distribution shift on retrieval performance, we analyze it through the lens of domain adaptation theory~\cite{redko2020survey}. Let $\mathcal{P}_{\mathrm{seen}}$ denote the distribution of seen tools, and $\mathcal{P}_{\mathrm{unseen}}$ denote the distribution of unseen tools. For any retrieval function $g(\cdot)$, the retrieval error on unseen tools is bounded by:
\begin{equation} 
\epsilon_{\mathrm{unseen}}(g(\cdot)) \leq \epsilon_{\mathrm{seen}}(g(\cdot)) + \operatorname{KL}(\mathcal{P}_{\mathrm{unseen}} \| \mathcal{P}_{\mathrm{seen}}), \label{eq:domain_adaptation} 
\end{equation} 
where $\epsilon_{\mathrm{unseen}}(g(\cdot))$ represents the expected retrieval error on unseen tools; $\epsilon_{\mathrm{seen}}(g(\cdot))$ represents the retrieval error on seen tools; and $\operatorname{KL}(\mathcal{P}_{\mathrm{unseen}} \| \mathcal{P}_{\mathrm{seen}})$ quantifies the KL divergence between the distributions of unseen and seen tools, measuring the degree of distribution shift. This reveals that directly encoding unseen tool representations using models trained on seen tools enlarges the error bound.  Consequently, this results in biased representations fail to capture the true functionality of unseen tools, leading to retrieval failures in the inductive setting.

\textbf{Challenge \ding{183}: Sensitivity of Semantic-only Retrieval.} Traditional retrieval-based methods primarily rely on computing semantic similarity between instructions and tool descriptions. However, this approach is highly sensitive to the quality of learned representations. When dealing with unseen tools, their representations often suffer from distribution shifts, causing semantic similarity calculations to produce severely misleading results.

\begin{table}[t]
	\centering
\caption{Retrieval error case illustrating limitations of the semantic-only method on unseen tools.}
	\small
    \vspace{-2mm}
	\begin{tabular}{p{0.95\columnwidth}}
		\toprule
		\textbf{Instruction:} I'm planning a family gathering and I want to make some non-alcoholic cocktails for the kids. Can you suggest some fun and colorful mocktail recipes? Also, provide me with the contact details of a brewery that offers non-alcoholic beer options. \\
		\midrule
		\textbf{Semantic-only retrieval:} [Food, Tasty, recipes/detail] \\
		\midrule
		\textbf{Label:} [Food, The Cocktail DB, Detailed Cocktail Recipe by ID] \\
		\bottomrule
	\end{tabular}
    \vspace{-3.8mm}
    \label{tab:sensitive}
\end{table}

From the example in Table~\ref{tab:sensitive}, consider the unseen tool ``The Cocktail DB'' in ToolBench. Given an instruction about ``mocktail recipes'' for a family gathering, this tool should ideally be retrieved, as it specializes in cocktail recipe services. However, because ``The Cocktail DB'' is an unseen tool not present in the training data, its representation suffers from distribution shift, causing it to be incorrectly positioned in the embedding space. As a result, semantic similarity calculations on this biased representation retrieve the ``recipes/detail'' tool, which matches the ``recipes'' keyword but lacks cocktail-specific functionality.

% As shown in Figure \ref{fig:pre_exp} (b), we measured the KL divergence between the representation distributions of unseen tools \textit{v.s.} existing tools in the training data, as well as between their corresponding instructions. We observe that both unseen instructions and unseen tools exhibit distribution shifts from training data, with unseen tools demonstrating significantly larger distributional divergence. This is because instructions, which typically follow common linguistic patterns and semantic structures, remain relatively stable across different tasks. In contrast, tools exhibit substantial functional diversity, as each tool may have distinct capabilities, input/output formats, and parameter configurations that vary greatly across domains. Consequently, directly encoding unseen tool representations tends to introduce bias that fails to capture their true functionality. In the inductive setting, models must handle both unseen instructions and unseen tools simultaneously, leading to compounded errors that severely degrade retrieval performance.

\subsection{Discussion and Insights}

As analyzed above, traditional methods rely only on semantic information, making them sensitive when handling unseen tools. These insights inform our revisit of the retrieval paradigm. Beyond semantic information, we need to identify other potential information.

Drawing from practical experience, we notice two intuitive patterns in tool usage: (i) Each tool tends to work with a relatively fixed and small set of companion tools. For example, when users book a flight, they typically need flight search, price comparison, booking, and payment tools working together. A payment tool consistently appears alongside booking tools, but rarely with unrelated tools like weather forecasting. (ii) When users express similar needs using different phrasings, such as "check my account balance" versus "view my current funds," they essentially require the same set of tools despite the semantic variations.

\begin{figure*}[t]
    \centering   
    \includegraphics[width=0.98\textwidth]{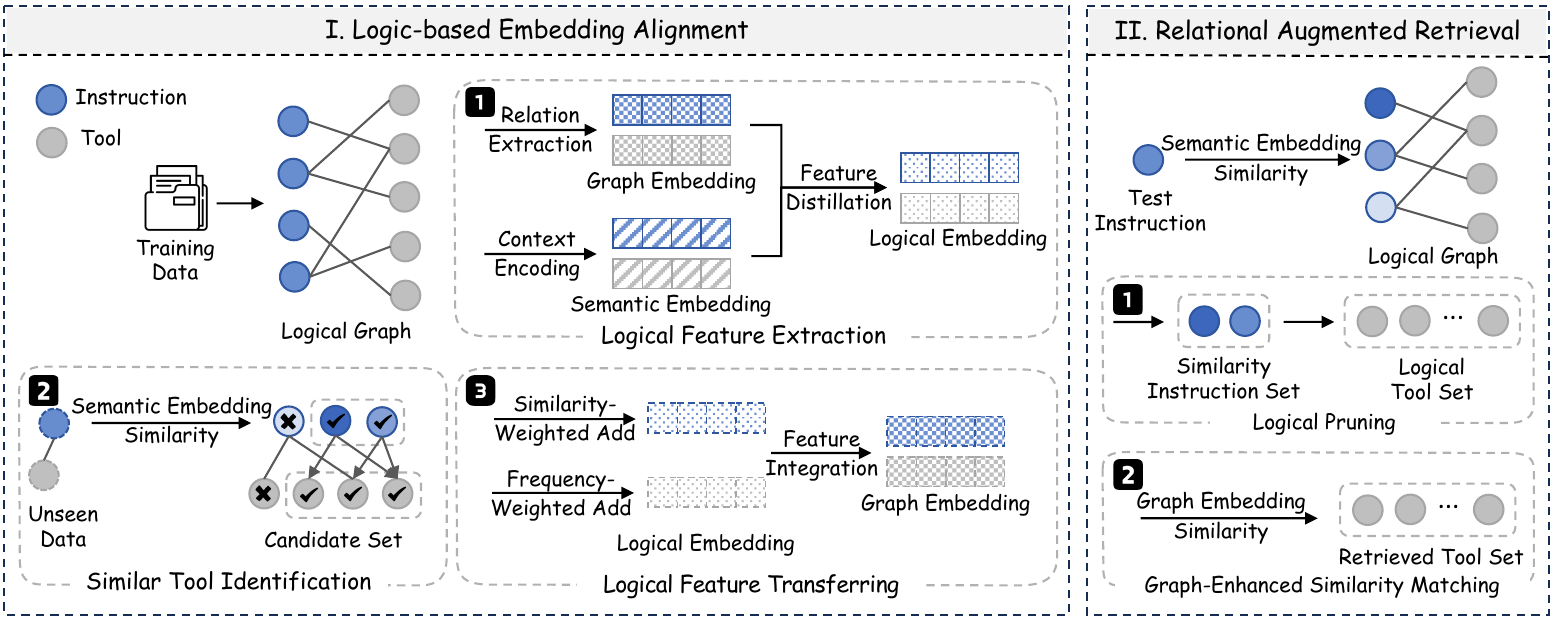}    
\caption{\textbf{The overall framework of \texttt{LoSemB}.} }
     \label{fig:framework}
     \vspace{-0.3cm}
\end{figure*}

To validate these intuitions, we conduct an empirical analysis on the I2 and I3 subsets of the ToolBench dataset. As shown in Figure \ref{fig:pre_exp} (c), each tool maintains a limited set of co-occurrence partners. Specifically, in the ToolBench (I2) dataset, 58.2\% of tools co-occur with fewer than 10 other tools, while an additional 24.6\% co-occur with only 10-20 tools. And, refer to Figure \ref{fig:pre_exp} (d), for each instruction, we identify the Top-5 similar instructions and find that, on average, 85.7\% of tools corresponding to each instruction overlap with the tool set associated with its similar instructions. In conclusion, we can leverage logical information from tool relationships and usage patterns to learn more accurate representations for unseen tools and overcome the sensitivity of semantic-only retrieval.

% These observations suggest that tool relationships and usage patterns contain valuable structural information beyond semantic similarity.
% In conclusion, retrieval based purely on textual semantics overlooks logical information, \textit{i.e.}, tool co-occurrence relationships and instruction-tool invocation patterns, resulting in vulnerability when handling unseen tool scenario.

\section{LoSemB Framework}
\subsection{Overview}

Based on our preliminary findings, logical information serves as a crucial component for addressing the challenges in inductive tool retrieval. Accordingly, we design a two-stage framework comprising offline indexing and online retrieval, as illustrated in Figure \ref{fig:framework}.  In the offline indexing phase, we first construct a graph containing tool nodes and instruction nodes. Building upon this graph, we introduce a logic-based embedding alignment module that extracts latent logical features and transfers them to unseen tools. In the online retrieval phase, we implement a relational augmented retrieval module. This module leverages the encoded logical relationships from the graph to guide the retrieval process by combining logical pruning with enhanced similarity matching, ultimately identifying the most relevant tools.

% 我们在前面的Preliminary findings 我们可以利用tool co-occurrence relationships and instruction-tool 逻辑 patterns,Thus, 我们需要在framework中解决两个问题: ❶ How could we leverage logical information, specifically extracting86the hidden logical features and transferring them to eliminate distribution shifts and learn better87representations for unseen tools without costly retraining? And ❷ how could we effectively integrate88logical information for more accurate retrieval, rather than relying only on text similarity, to guide a89more robust and accurate tool retrieval process?息，从人类学习使用unseen tool中获得灵感，我们首先要通过过往的经验，学习新工具的用途，并且再在过往经验的指导下，实践使用新的工具，Motivated by this idea, 我们的离线索引阶段类似于记忆编码, we initially introduce a new tool retrival paradigm in Figure \ref{fig:framework}. It constructs a graph that contains two types of nodes: entity nodes and instruction nodes. 然后我们需要把记忆嵌入到unseen tool which employs a logic-based embedding alignment module that157extracts and utilizes logical features to eliminate distribution shifts，最后 implements relational158augmented retrieval mechanism hat further enhances performance through combining logical159constraints and graph-enhanced similarity matchin

\subsection{Logical Graph Construction}

Given a corpus with a set of $n$ instructions $\mathcal{Q}=\{q_1,\ldots, q_n\}$ and a tool repository $\mathcal{T}=\{t_1,\ldots, t_m\}$ containing $m$ tools, all of which are available during retriever training. Then, we construct a bipartite graph, called the logical graph, denoted as $\mathcal{G}=(\mathcal{V}, \mathcal{E}, \mathcal{D})$, where $\mathcal{V}=\mathcal{Q} \cup \mathcal{T}$ and $\mathcal{D}$ contains the textual descriptions associated with the nodes in $\mathcal{V}$. 

The edges $\mathcal{E}$ are constructed according to the following rule: if an instruction $q_i$ requires a tool $t_j$ to satisfy its request, we add an edge $(v_{q_i}, v_{t_j})$. These relations correspond to an incidence matrix, called the \textit{invoke matrix} $\mathbf{R}$, between instructions and tools.

Formally, the invoke matrix $\mathbf{R}$ is defined as an $n \times m$ matrix:
\begin{equation}
    \mathbf{R} = [r_{ij}]_{n \times m};\;
    r_{ij} = \mathds{1}(q_i \leftrightarrow t_j). 
\end{equation}
Here, $r$ is the entry of matrix $\mathbf{R}$; $\leftrightarrow$ represents a bidirectional  invocation relation from instruction to tool; $\mathds{1}$ denotes the indicator function, where $r_{ij}=1$ if instruction $q_i$ invokes tool $t_j$ ($q_i \leftrightarrow t_j$), and $r_{ij}=0$ if there is not invocation relation. 

We encode the textual content of instructions and the descriptions of tools to generate initial embeddings using a fine-tuned pre-trained language model (PLM). Specifically, we obtain the instruction embeddings $\mathbf{E}^{(0)}_q \in \mathbb{R}^{n \times d}$ and the tool embeddings $\mathbf{E}^{(0)}_t \in \mathbb{R}^{m \times d}$.  We refer to these initial embeddings as semantic embeddings. Then we concatenate them as $\mathbf{E}^{(0)} = \mathbf{E}^{(0)}_q \oplus \mathbf{E}^{(0)}_t \in \mathbb{R}^{(n+m) \times d}$, where $d$ denotes the embedding dimension.

\subsection{Logic-based Embedding Alignment}

Our analysis in Section \ref{sec:kl} reveals that unseen tools suffer from the large distribution shift, causing degraded retrieval performance. Our key insight is that when tools have similar functions, they will demonstrate similar instruction-tool invoke patterns and tool co-occurrence relationships in the graph. Therefore, based on this logical information, we can extract logical features from functionally similar seen tools and transfer them to unseen tools.

However, it raises several critical questions, \textit{i.e.}, how to extract latent logical features, how to select nodes from which to transfer features, and how to transfer them to unseen nodes. We address these through three stages: logical feature extraction, similar tool identification, and logical feature transfer.

\textbf{Logical Feature Extraction.} We first need to extract logical features from existing nodes. Our approach extracts logical features by capturing how graph convolution transforms the original semantic embeddings by incorporating the logical information of the instruction-tool invoke pattern and tool co-occurrence relationships. As a result, during graph convolution, nodes change from their semantic embeddings, and then we can capture the logical features by finding the difference between the graph embeddings and the original embeddings.

% Specifically, we employ a multi-layer graph convolutional mechanism, following LightGCN~\cite{he2020lightgcn}, 
% \begin{equation}
% {E}^{(k+1)} = {D}^{-\frac{1}{2}} {A} {D}^{-\frac{1}{2}} {E}^{(k)},
% \end{equation}
% where ${A} = \begin{bmatrix} {0}_{N \times N} & \mathcal{R} \\ \mathcal{R}^T & {0}_{M \times M} \end{bmatrix}$ is the adjacency matrix constructed from the invoke matrix $\mathcal{R}$, and ${D}$ is the diagonal degree matrix with $d_{ii} = \sum_{j} a_{ij}$.

% After $K$ layers of propagation, we merge embeddings from all layers to form the final graph embeddings:
% \begin{equation}
% E = \frac{1}{K+1} \sum_{k=0}^{K} E^{(k)}
% \end{equation}

% We then perform feature distillation through a comparative transformation that maps graph embeddings against original text embeddings to obtain logical embeddings:
% \begin{equation}
% \Delta {E} = E-E^{(0)}.
% \end{equation}

% Here, $\Delta$ represents the logical feature matrix, where $\Delta {E}_{t_j}$ and $\Delta {E}_{q_i}$ correspond to the logical features of tool and instruction nodes. Thus, this approach provides three complementary representations for each node: text embeddings contain semantic content, graph embeddings combine logical and semantic content, and logical features capture the node's functional role within the graph.
Specifically, we employ a multi-layer graph convolutional mechanism, following LightGCN~\cite{he2020lightgcn}, 
\begin{equation}
\mathbf{E}^{(l+1)} = \mathbf{D}^{-\frac{1}{2}} \mathbf{A} \mathbf{D}^{-\frac{1}{2}} \mathbf{E}^{(l)},
\end{equation}
where $\textbf{A} = \begin{bmatrix} {0}_{n \times n} & \mathbf{R} \\ \mathbf{R}^T & {0}_{m \times m} \end{bmatrix}$ is the adjacency matrix constructed from the invoke matrix $\mathbf{R}$, and $\mathbf{D}$ is the diagonal degree matrix with $d_{ii} = \sum_{j} a_{ij}$, where $a_{ij}$ is the entry of $\mathbf{A}$.

After $L$ layers of propagation, we merge embeddings from all layers to form the final graph embeddings:
\begin{equation}
\mathbf{E} = \frac{1}{L+1} \sum_{l=0}^{L} \mathbf{E}^{(l)}
\end{equation}

We then perform feature distillation through a comparative transformation that maps graph embeddings against original semantic embeddings to obtain logical embeddings:
\begin{equation}
\Delta \mathbf{E} = \mathbf{E}-\mathbf{E}^{(0)}.
\end{equation}

Here, $\Delta$ represents the logical feature matrix, where $\Delta \mathbf{E}_{t_j}$ and $\Delta \mathbf{E}_{q_i}$ correspond to the logical features of tool and instruction nodes. Thus, this approach provides three complementary representations for each node: semantic embeddings contain content information, graph embeddings combine logical and semantic content, and logical features capture the node's functional role within the graph.

\textbf{Similar Tool Identification via Structured Propagation.} After extracting logical features, we need to identify which existing nodes contain the most similar logical features for unseen tools. While intuitively we might search for textually similar tools as sources for knowledge transfer, this approach performs poorly when the retriever struggles to generate accurate representations for unseen tools. Moreover, textually similar tools do not necessarily share functional similarities. Our key insight is that tools that process similar instructions are more likely to share similar functions and logical features. Therefore, we use the unseen tool's associated instruction as a bridge to locate functionally similar tools through a structured propagation mechanism.

Formally, given a set of unseen tools $\mathcal{T}^u = \{t^{u}_{1}, \ldots, t^{u}_{h}\}$, for each unseen tool $t^{u}$ with its associated instruction $q^{u}$, we first encode them using the same PLM to obtain their initial semantic embeddings $\mathbf{e}_{t^{u}}^{(0)}$ and $\mathbf{e}_{q^{u}}^{(0)}$. We then compute the contextual association between the unseen instruction $q^{u}$ and all training instructions $\mathcal{Q}$ based on their semantic embeddings, representing these similarities as a vector:
\begin{equation}
\mathbf{s} = [s_{i}]_{n \times 1};\;s_{i} = \cos(\mathbf{e}_{q_i}^{(0)}, \mathbf{e}_{q^{u}}^{(0)}). 
\end{equation}
This vector $\mathbf{s}$ captures how similar each training instruction $q \in \mathcal{Q}$ is to the unseen instruction $q^{u}$, providing a comprehensive similarity distribution across the entire training instruction space. To identify the most relevant instruction candidates while filtering out noise, we apply a top-$J$ selection mechanism that constructs a sparse activation vector $\mathbf{c}= [c_{i}]_{n \times 1}$, where $c_{i} = \mathds{1}(i\in\operatorname{Top}\!-\!J(s)) \cdot s_{i}$. Here $\mathds{1}$ is the indicator function and $\operatorname{Top}\!-\!X(\cdot)$ operator returns the indices of the top $X$ elements of the input set. We then propagate this sparse instruction activation through the invoke matrix $\mathbf{R}$ to aggregate activation scores for candidate tools via matrix multiplication:
\begin{equation}
\mathbf{b} = \mathbf{R}^T \mathbf{c} \in \mathbb{R}^{m \times 1},
\end{equation}
where $\mathbf{b} = [b_{j}]_{m \times 1}$ represents the aggregated activation scores for all training tools $\mathcal{T}$. Each element $b_{j}$ quantifies the functional similarity between unseen tool $t^{u}$ and training tool $t_j$, computed by aggregating the activation weights from all similar instructions that invoke $t_j$. Intuitively, if training tool $t_j$ is invoked by some instructions similar to $q^{u}$, it receives a higher activation score, suggesting that $t_j$ serves similar functional roles as the unseen tool $t^{u}$. We identify tools with positive activation scores as functionally similar candidates: $\mathcal{T}_\mathrm{cand} = \{t_j | b_{j} > 0, t_j \in \mathcal{T}\}$.

\begin{table*}[t]
\caption{\textbf{Results (\%) of baselines and \texttt{LoSemB} in transductive and inductive settings across five datasets.} We report the average performance across R@3, R@7, P@3, and P@7. The best result for each dataset is highlighted in \textbf{bold}, and the second-best is indicated with an \underline{underline}.}
    \label{tab:main_results}
    \small
    \centering
    % \vspace{-3mm}
    \setlength{\tabcolsep}{1mm}
    \scalebox{0.96}{
    \begin{tabular}{lcccccccccc}
    \toprule
    \multirow{2}{*}{\textbf{Method}}
        &\multicolumn{2}{c}{\textbf{ToolBench (I1)}}
        & \multicolumn{2}{c}{\textbf{ToolBench (I2)}}
        &\multicolumn{2}{c}{\textbf{ToolBench (I3)}}
        &\multicolumn{2}{c}{\textbf{UtralTool}}
        &\multicolumn{2}{c}{\textbf{APIGen}}\\
        \cmidrule(lr){2-3} \cmidrule(lr){4-5} \cmidrule(lr){6-7} \cmidrule(lr){8-9} \cmidrule(lr){10-11}
       &Transductive &Inductive &Transductive &Inductive  &Transductive &Inductive  &Transductive &Inductive  &Transductive &Inductive \\
    \midrule
    \multicolumn{11}{c}{\textbf{\textit{Direct Zero-shot LLM Inference}}} \\
    \midrule
        GPT-3.5-turbo &1.31 &1.31 &1.91 &1.91 &10.79 &10.79 &6.51 &6.51 &3.95 &3.95 \\
        GPT-4o-mini &10.45 &10.45 & 8.67& 8.67&48.26 & 48.26& 29.55& 29.55&18.50 &18.50 \\
    \midrule
    \multicolumn{11}{c}{\textbf{\textit{Training-free Tool Retrieval Method}}} \\
    \midrule
        BM25 &17.37 &17.37 &13.29 &13.29 &15.77 &15.77 & 9.22& 9.22&34.60 &34.60 \\
        Ada Embedding &45.90 &45.90 &26.34 &26.34 &34.75 &34.75 &26.56 &26.56&39.00 &39.00\\
        all-MiniLM-L6-v2 & 32.75& 32.75& 17.76& 17.76& 23.90&23.90 &19.45 & 19.45&34.40 &34.40 \\
        Re-invoke &38.06 &38.06 &30.69 &30.69 &37.25 &37.25 &14.17 &14.17 &32.00 &32.00 \\
    \midrule
    \multicolumn{11}{c}{\textbf{\textit{Training-based Tool Retrieval Methods}}} \\
    \midrule
        EmbSim &64.56 & 58.75& 53.99& 44.35& 58.86&46.09 &35.53 &26.87 &45.65 &43.60 \\
        InstRet &\underline{66.35} &\underline{61.93} &\underline{56.39} &44.70 &60.68 &48.62 &36.11 &29.54&46.00 &44.00 \\
        RerankRet  & 65.78& 60.74& 53.77&\underline{50.15} &\underline{68.03} &\underline{54.35} &\underline{61.74} &\underline{48.83}&\underline{48.85} &\underline{46.20}\\
        \midrule
        \rowcolor[HTML]{cbddf5} \textbf{LoSemB (Ours)} &\textbf{70.46} &\textbf{68.44} &\textbf{66.37} &\textbf{64.41} &\textbf{71.47} &\textbf{68.48} &\textbf{67.55} &\textbf{65.35} &\textbf{60.20} & \textbf{58.45}\\
    \bottomrule
    \end{tabular}}
    % \vspace{-2mm}
\end{table*}

\textbf{Logical Feature Transfer.} After identifying candidate nodes, we design tailored weighting mechanisms to prioritize the most valuable logical features from candidate nodes. For unseen tools, we employ a frequency-based weighting strategy motivated by the insight that tools frequently invoked by similar instructions are likely to share similar logical patterns. Specifically, the weight for each candidate tool $t_j \in \mathcal{T}_\mathrm{cand}$ is computed by normalizing its invocation frequency among the top-$J$ similar instructions:
\begin{equation}
w_{j} = \frac{\sum_{i=1}^{n} r_{ij} \cdot \mathds{1}_{\{c_{i}>0\}}}{\sum_{t_p \in \mathcal{T}_\mathrm{cand}} \sum_{i=1}^{n} r_{ip} \cdot \mathds{1}_{\{c_{i}>0\}}},
\end{equation}
where $r_{ij}$ is the entry of invoke matrix, and $\mathds{1}_{\{c_{i}>0\}}$ indicates whether instruction $q_i$ is among the top-$J$ similar instructions. Using these weights, we apply feature integration to generate the graph embedding for the unseen tool $t^{u}$ by combining its semantic embedding with the weighted logical features:
\begin{equation}
\mathbf{e}_{t^{u}} = \mathbf{e}_{t^{u}}^{(0)} + \sum_{t_j \in \mathcal{T}_\mathrm{cand}} w_{j} \cdot \Delta \mathbf{e}_{t_j}.
\end{equation}

For unseen instructions, we adopt a similarity-based softmax normalization strategy to ensure more semantically similar instructions contribute more significantly to the feature transfer. The weight for each selected training instruction is calculated as:
\begin{equation}
w_{i} = \frac{\exp(s_{i})}{\sum_{q_i : c_{i}>0} \exp(s_{i})},
\end{equation}
where $s_{i}$ is the cosine similarity and the denominator only includes instructions with non-zero activation. We then generate the graph embedding for the unseen instruction $q^{u}$ with the weighted logical features:
\begin{equation}
\mathbf{e}_{q^{u}} = \mathbf{e}_{q^{u}}^{(0)} + \sum_{q_i : c_{i}>0} w_{i} \cdot \Delta \mathbf{e}_{q_i}.
\end{equation}

Finally, by aligning unseen nodes with existing nodes through weighted logical feature transfer, we obtain better representations of unseen tools without requiring costly retraining, and then seamlessly integrate them into the logical graph $\mathcal{G}$.

\subsection{Relational Augmented Retrieval}
\label{sec:relational retrieval}

% Traditional retrieval methods only rely on computing semantic similarity between instructions and tool descriptions, making them extremely sensitive to representation quality. This limitation becomes particularly severe in inductive settings, where models must handle entirely unseen tools during inference. 

% Following on the graph-enhanced embeddings, we 进一步 propose a relational augmented retrieval mechanism relies on observed instruction-tool relationships rather than solely on semantic representations of tools, it is less sensitive to representation quality. 从而一方面这个鲁棒性提高另一方面我本身就利用了更好的mebdding 

Our key insight is if certain tools are frequently invoked by similar training instructions, they are likely relevant to semantically related test instructions. Building on this observation, we leverage the logical graph to exploit such relational patterns. Since our approach relies on observed instruction-tool relationships rather than solely on semantic representations of tools, it is less sensitive to representation quality. Furthermore, we refine similarity matching using graph-enhanced embeddings that incorporate logical information.

% Traditional retrieval methods only rely on computing semantic similarity between instructions and tool descriptions, making them extremely sensitive to representation quality. This limitation becomes particularly severe in inductive settings, where models must handle entirely unseen tools during inference. To address this critical limitation, we propose a relational augmented retrieval mechanism that leverages logical graph structures to prune the search space and employs graph-enhanced embeddings to refine the similarity matching process.

\textbf{Logical Pruning.} Given a test instruction $q^{\mathrm{test}}$, we first compute its similarity distribution with all training instructions:

\begin{equation}
\mathbf{s}^{\mathrm{test}} = [s^{\mathrm{test}}_{i}]_{n \times 1};\;s^{\mathrm{test}}_{i} = \cos(\mathbf{e}_{q_i}^{(0)}, \mathbf{e}_{q^{\mathrm{test}}}^{(0)}),
\end{equation}
where $q_i \in \mathcal{Q}$ is the training instruction set. We then identify the top-$I$ most similar instructions and collect all tools invoked by them as the pruned candidate set:
\begin{equation}
\mathcal{T}_{\mathrm{logic}} = \{t_j | (q_i, t_j) \in \mathcal{E}, i \in \operatorname{Top}\!-\!I(s^{\mathrm{test}})\}.
\end{equation}
This logic-based pruning effectively reduces the search space from $m$ tools to $|\mathcal{T}_{\mathrm{logic}}| \ll m$ by focusing on tools that have demonstrated usefulness for similar instructions, leveraging the structural patterns captured in the logical graph $\mathcal{G}$.

\textbf{Graph-Enhanced Similarity Matching.} Within the pruned candidate set $\mathcal{T}_{\mathrm{logic}}$, we perform graph-enhanced similarity matching that incorporates both semantic and logical information to achieve more accurate similarity matching. We first generate the graph embedding for the test instruction $\mathbf{e}_{q^{\mathrm{test}}}$ following the same feature transfer process described earlier. (Note that during this process, we fuse the logical features from the top-$I(s)$ most similar instructions in the training set.). Then, we retrieve the top-$K$ most relevant tools by ranking the cosine similarity between the test instruction's graph embedding and each candidate tool's graph embedding. The retrieval process is formulated as:
\begin{equation}
\mathcal{T}_{\mathrm{ret}}' = \{ t_j \mid t_j \in \mathcal{T}_{\mathrm{logic}},\ j \in \operatorname{Top}\!-\!K(\cos(\mathbf{e}_{q^{\mathrm{test}}},\mathbf{e}_{t_j})) \}
\end{equation}

In summary, this two-stage retrieval mechanism synergistically integrates coarse-grained logical pruning and fine-grained graph-enhanced similarity matching, enabling both efficient and accurate tool retrieval even for unseen tools.

% full tool table

\section{Experiment}
In this section, we conduct comprehensive experiments to verify the effectiveness and efficiency of \texttt{LoSemB}. Specifically, we aim to answer the following questions. \textbf{Q1: (Retrieval Quality):} How does \texttt{LoSemB} perform in terms of tool retrieval accuracy compared with state-of-the-art tool retrieval methods across both transductive and inductive settings? \textbf{Q2: (Generation Performance):} How does \texttt{LoSemB} perform in downstream generation tasks when compared against state-of-the-art tool retrieval methods? \textbf{Q3 (Ablation Studies):} What contribution does each key component of \texttt{LoSemB} make to the overall performance? \textbf{Q4 (Noise Analysis):} How robust is \texttt{LoSemB} to incorrect instruction-tool pairs in the training data compared to baseline methods? (Note that for an in-depth evaluation of \texttt{LoSemB}, additional experiments on efficiency analysis, backbone analysis, parameter sensitivity and case study are presented in Appendix \ref{app:add_exp}.)

\subsection{Experimental Setup}
\label{sec:set+}

\textbf{Datasets.} We evaluate the performance of \texttt{LoSemB} on three tool learning benchmarks: ToolBench~\citep{qintoolllm}, UltraTool~\citep{huang2024planning}, and APIGen~\citep{lin2024hammer}. Specifically, ToolBench is a large-scale synthetic benchmark containing over 16,000 tool collections with approximately 47,000 unique API interfaces across three distinct scenarios. APIGen provides 3,673 executable APIs spanning 21 categories. In contrast to these synthetic benchmarks. UltraTool comprises 5,824 real-world instruction examples across 22 diverse domains, incorporating 2,032 tools. More details can be found in Appendix \ref{app:data}.

% ToolBench~\cite{qintoolllm}, a tool benchmark containing over 16k tool collections, each containing several APIs, totaling about 47,000 unique interfaces. For simplicity, we refer to each API as a tool in this paper. Our evaluation follows the established data categorization, which spans three distinct scenarios: single-tool instructions (I1), intra-category multi-tool instructions (I2), and intra-collection multi-tool instructions (I3). More details can be found in Appendix \ref{app:toolbench}, and we also conduct experiments on the UltraTool~\cite{huang2024planning} dataset in Appendix \ref{app:ultra}. We follow the same evaluation method a

\textbf{Baselines.} We categorize all the baselines into three groups: (i) Zero-shot LLM Inference. We evaluate several foundational models, including GPT-3.5-turbo and GPT-4o-mini~\citep{openai2023gpt4}. Given the different input constraints of LLMs, we select 400 and 2,000 tools, respectively, to incorporate into the prompt, where the LLM directly selects the relevant tools. (ii) Training-free Tool Retrieval Methods. We compare against several retrieval baselines that require no domain-specific training: BM25~\cite{robertson2009probabilistic}, a sparse retrieval method based on lexical matching; Ada Embedding, a dense retrieval approach that uses OpenAI's text-embedding-ada-002 model to encode instructions and tool documents for similarity matching; all-MiniLM-L6-v2~\citep{wang2020minilm}, a pre-trained sentence-transformers model that computes semantic similarity between instructions and tools; and Re-Invoke~\cite{chen2024re}, an unsupervised baseline that enhances retrieval through query rewriting and document expansion to bridge the semantic gap between instructions and tool descriptions. (iii) Training-based Tool Retrieval Methods. We compare against several methods that require domain-specific training on instruction-tool pairs: EmbSim, a retriever directly fine-tuned on instruction-tool pairs to learn task-specific embeddings; InstRet (Instruction-Augmented Retrieval), inspired by Re-Invoke, which enriches the training data by integrating tool documents with instructions to create augmented instruction-tool pairs for fine-tuning; and RerankRet (Reranking-Augmented Retrieval), a two-stage retrieval pipeline that first retrieves candidate tools using fine-tuned embeddings and then applies a cross-encoder reranker for final selection.

\textbf{Evaluation Metrics.} For retrieval quality assessment, we use \textit{Recall@$K$} and \textit{Precision@$K$} with $K \in \{3, 7\}$. We do not include nDCG as it is not well-suited for tool retrieval scenarios~\cite{moon2024efficient,wang2013theoretical}. In tool retrieval tasks, tool relevance is binary, and the order of retrieved tools does not affect performance. For downstream generation performance, we adopt metrics from StableToolBench~\citep{guo2024stabletoolbench}, including \textit{Pass Rate}, which measures whether the model successfully completes a given task, and \textit{Win Rate}, which employs an LLM judge to determine which approach produces superior results. Here, we conduct comparisons between outputs generated using baseline-retrieved tools and those generated using \texttt{LoSemB}-retrieved tools.

\textbf{Implementation Details.} We conduct experiments under two settings. In the transductive setting, we filter the test set to remove instructions involving tools not present in the training data, ensuring all tools are included in the training set. In the inductive setting, we randomly select 10\%, 20\%, and 30\% of test tools as unseen tools and exclude them from the training data. Within \texttt{LoSemB}, retrievers are trained using BERT-base as the backbone model.
% three backbone models with varying architectures and sizes: all-miniLM-L6-v2 (22.7M parameters)\cite{wang2020minilm}, referred to as miniLM for short, DeBERTaV3-base (86M parameters)\cite{he2021debertav3}, and BERT-base (110M parameters)~\cite{devlin2019bert}. More details, including \texttt{LoSemB} parameter settings, can be seen in Appendix \ref{app:exp_details}.

% \textbf{Metrics.} We evaluate performance using Recall@$K$ and Precision@$K$, with $K$ values of 3 and 7. We do not include nDCG as it is not well-suited for tool retrieval scenarios~\cite{moon2024efficient,wang2013theoretical}, unlike ToolRetriever~\cite{qintoolllm} and Re-invoke~\cite{chen2024re}, which use this metric. In tool retrieval tasks, tool relevance is binary, and the order of retrieved tools is not important.
% \begin{figure*}[t]
%     \centering   
%     \includegraphics[width=\textwidth]{figure/relitu.pdf}     
%      \caption{\textbf{Results(\%) of different tool retrieval baselines using the BERT-base and all-miniLM-L6-v2 backbones in the inductive setting.} The x-axis represents the percentage of unseen tools,  while the y-axis denotes the different baselines. Each cell contains two values: the absolute performance score, and "\textcolor{gray}{$\downarrow$}" denotes the relative performance drop compared to the 0\% unseen tool scenario. We see that \texttt{LoSemB} shows much smaller performance degradation compared to other baselines.}
%      \label{fig:ins}
% \end{figure*}   
\subsection{Retrieval Quality (Q1)} 

\begin{figure}[t]
  \centering
  \includegraphics[width=0.99\linewidth]{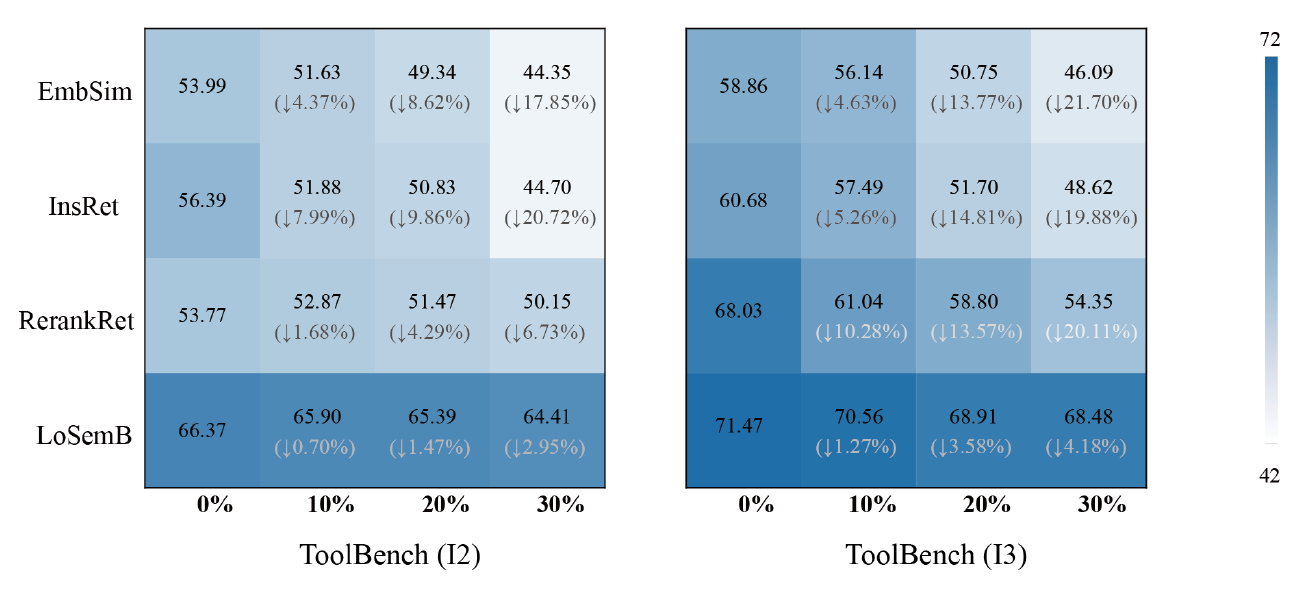}
  \vspace{-4mm}
\caption{Retrieval results (\%) of different baselines across varying percentages of unseen tools. Each cell shows the absolute performance and the relative drop (``$\downarrow$'') compared to the transductive setting.}
\vspace{-4mm}
\label{fig:relitu}
\end{figure}

To address Q1, we conduct an evaluation of retrieval accuracy by comparing various baseline methods with \texttt{LoSemB} across five benchmark datasets. Table \ref{tab:main_results} evaluates performance under both transductive and inductive settings, where the transductive setting contains no unseen tools in the test set, while the inductive setting includes 30\% unseen tools. Figure \ref{fig:relitu} further illustrates the performance degradation trends of training-based tool retrieval methods across varying unseen tool ratios. Based on these experimental results, we derive the following key observations.

\textbf{Obs. 1. Tool retrieval methods enhance zero-shot LLM performance.} Direct prompting of LLMs with a partial tool repository, constrained by limited context windows, yields the poorest performance across most datasets, except when the tool repository is small enough to fit entirely within the context. For instance, GPT-4o-mini achieves only 8.67\% accuracy on ToolBench(I2). In contrast, incorporating a retriever to search for the most relevant tools from the large repository significantly improves performance. However, its effectiveness may vary across different retrievers. Specifically, while lexical methods like BM25 show unstable results, pre-trained dense embedding models such as Ada Embedding deliver consistent improvements, boosting accuracy to 26.34\% on ToolBench (I2).

\textbf{Obs. 2. Domain-specific training aligns user intent with tool descriptions better than general-purpose retrievers, but exhibits limited generalization to unseen tools.} Training-based methods consistently outperform training-free approaches, demonstrating the benefits of domain-specific training. However, when moving to the inductive setting with unseen tools, all trained methods exhibit performance degradation across all datasets. For instance, RerankRet achieves solid results on most datasets in the transductive setting, reaching 61.74\% on UltraTool, but suffers a significant drop to 48.83\% in the inductive setting.

\textbf{Obs. 3. \texttt{LoSemB} demonstrates superior performance compared to all baselines in both transductive and inductive settings.} While training-free methods maintain stable performance across both transductive and inductive settings, they achieve suboptimal results. Typical training-based retrieval methods also struggle with frequently updated tool repositories, as maintaining high performance typically requires costly retraining. In contrast, by leveraging logical information, \texttt{LoSemB} outperforms all baseline methods by a significant margin across all datasets. Furthermore, as shown in Figure \ref{fig:relitu}, \texttt{LoSemB} exhibits more stable performance as the proportion of unseen tools increases. Specifically, on ToolBench (I3), \texttt{LoSemB} maintains stable performance with relative decreases of merely 1.27\%, 3.58\%, and 4.18\%, respectively.

\subsection{Generation Performance (Q2)}

\begin{table}[h]
\vspace{-2mm}
\centering
\caption{Generation results (\%) of different baselines. The table compares the pass rate and win rate for transductive and inductive settings across ToolBench (I2) and ToolBench(I3).}
\vspace{-2mm}
\resizebox{0.45\textwidth}{!}{%
\begin{tabular}{c|l|cc|cc}
\toprule
\multirow{2}{*}{Dataset} & \multirow{2}{*}{Method} & \multicolumn{2}{c|}{Transductive} & \multicolumn{2}{c}{Inductive} \\
\cline{3-6}
& & Pass Rate & Win Rate & Pass Rate & Win Rate \\
\midrule
\multirow{5}{*}{I2} 
& Re-invoke & 37.25 & 41.37 & 37.25& 41.52 \\
& EmbSim & 44.73 & 43.84 & 42.68 & 42.83 \\
& InsRet & \underline{45.71} & 44.89 & 43.34 & 44.13 \\
& RerankRet & 43.06 & \underline{45.13} & 41.27 & \underline{44.78} \\
& LoSemB & \textbf{50.04} & NA & \textbf{48.92} & NA \\
\midrule
\multirow{5}{*}{I3} 
& Re-invoke & 40.87 & 40.19 & 40.87& 41.86 \\
& EmbSim & 48.16 & 45.79 & 45.23 & 44.65 \\
& InsRet & 49.15 & 47.44 & 45.08 & 42.79 \\
& RerankRet & \underline{53.72} & \underline{47.66} & \underline{47.38} & \underline{45.58} \\
& LoSemB & \textbf{57.24} & NA & \textbf{54.27} & NA \\
\bottomrule
\end{tabular}%
}
\label{tab:generation}
\vspace{-2mm}
\end{table}

% G2，G3 0，30两种情况，然后有两个指标，不同的baseline 柱状图
To investigate the impact of retrieval on generation, we use GPT-4o-mini as the task execution model to evaluate the generation quality of different tool retrieval methods under both transductive and inductive settings. As shown in Table \ref{tab:generation}, we observe that:

% adopt the experimental settings from StableToolBench to evaluate the generation quality of different tool retrieval methods under both transductive and inductive settings. We use GPT-4o-mini as the task execution model for this evaluation. From the results in Table \ref{tab:generation}.

\textbf{Obs. 4. LoSemB achieves superior generation quality with the highest task completion rate.} \texttt{LoSemB} consistently achieves the highest pass rate across both datasets and settings, reaching 50.04\% and 48.92\% on ToolBench(I2) in transductive and inductive settings, respectively. Furthermore, all baseline methods exhibit win rates below 50\% when compared against LoSemB, indicating that \texttt{LoSemB} produces higher-quality outputs than other baselines. This demonstrates that accurate tool retrieval translates directly into improved downstream task performance.

\subsection{Ablation Studies (Q3)}

We conduct systematic ablation studies on the core components of \texttt{LoSemB} on ToolBench (I2). We examine two key modules: \uppercase\expandafter{\romannumeral1.} Logic-based Embedding Alignment Module, with two variants including \textit{w/o Instruction Transferring (w/o Ins-Trans)}, which uses only semantic embeddings for test instructions to calculate similarity, and \textit{w/o Tool Transferring (w/o Tool-Trans)}, which directly adopts semantic embeddings for unseen tool nodes; and \uppercase\expandafter{\romannumeral2.}, Relational Augmented Retrieval Mechanism, \textit{i.e.}, \textit{w/o Relational Retrieval (w/o Rel-Retrieval)}, which computes similarity scores with all available tools. Each variant was evaluated across different ratios of unseen tools. From the experimental results in Table \ref{tab:ablation_study}, we can observe that:

\begin{table}[h]
\centering
\caption{\textbf{Ablation results (\%) on key modules of \texttt{LoSemB} on ToolBench (I2).} "Drop \textcolor[HTML]{3466fa}{$\downarrow$}" rows show relative performance degradation (\%) compared to the full model.}
\label{tab:ablation_study}
\resizebox{\columnwidth}{!}{%
\begin{tabular}{lcccccccc}
\toprule
\multirow{2}{*}{Variant} & \multicolumn{2}{c}{0\%} & \multicolumn{2}{c}{10\%} & \multicolumn{2}{c}{20\%} & \multicolumn{2}{c}{30\%} \\
\cmidrule(lr){2-3} \cmidrule(lr){4-5} \cmidrule(lr){6-7} \cmidrule(lr){8-9} 
 & R@3 & P@3 & R@3 & P@3 & R@3 & P@3 & R@3 & P@3 \\
\midrule
w/o Ins-Trans & 78.00 & 62.32 & 78.40 & 62.56 & 76.69 & 61.27 & 75.55 & 60.04 \\
Drop \textcolor[HTML]{3466fa}{$\downarrow$} & 3.12 & 3.20 & 1.54 & 1.48 & 2.34 & 2.16 & 2.14 & 2.18 \\
\midrule
w/o Tool-Trans & NA & NA & 78.57 & 62.73 & 76.96 & 61.56 & 75.58 & 59.98 \\
Drop \textcolor[HTML]{3466fa}{$\downarrow$} & NA & NA & 1.33 & 1.21 & 2.00 & 1.69 & 2.15 & 2.28 \\
\midrule
w/o Rel-Retrieval & 72.27 & 57.22 & 65.94 & 52.70 & 64.90 & 51.53 & 55.92 & 44.25 \\
Drop \textcolor[HTML]{3466fa}{$\downarrow$} & 10.23 & 11.12 & 17.19 & 17.01 & 17.36 & 17.71 & 27.60 & 27.91 \\
\midrule
Full model & \textbf{80.51} & \textbf{64.38} & \textbf{79.63} & \textbf{63.50} & \textbf{78.53} & \textbf{62.62} & \textbf{77.24} & \textbf{61.38} \\
\bottomrule
\end{tabular}}
% \vspace{-4mm}
\end{table}
% \vspace{-1mm}

\begin{table}[htbp]
\centering

\caption{KL divergence between unseen tool embeddings and training data before and after logic-based alignment}
\label{tab:kl_divergence_unseen}
\begin{tabular}{lcc}
\toprule
Variant & ToolBench (I2) & ToolBench (I3) \\
\midrule
w/ Logic-based Alignment & 0.001276 & 0.001132 \\
w/o Logic-based Alignment & 0.18183 & 0.14869 \\
\bottomrule
\end{tabular}
% \vspace{-2mm}
\label{tab:tool_kl}
\end{table}

\textbf{Obs. 5. Each module of \texttt{LoSemB} is critical for both transductive and inductive settings.} Our \texttt{LoSemB} achieves the best performance across different proportions of unseen tools. Notably, as the number of unseen tools increases, the contribution of our modules becomes larger, highlighting the crucial role of our approach in addressing inductive scenarios.

To further validate the effectiveness of Logic-based Embedding Alignment in addressing the distributional shift problem identified in Section \ref{sec:kl}. We measure the KL divergence between the embeddings of unseen tools \textit{v.s.} seen tools, comparing the distribution before and after applying the alignment mechanism.

\textbf{Obs. 6. Logic-based Embedding Alignment effectively resolves the distribution shift.} As shown in Table \ref{tab:tool_kl}, unseen tools initially exhibit a large distribution gap from seen data, with KL divergence values of 0.18183 (I2) and 0.14869 (I3). Logic-based Embedding Alignment reduces these to 0.001276 and 0.001132, respectively, aligning unseen tool embeddings with the distribution of seen data and enabling accurate retrieval by leveraging logical knowledge from seen data.

% 我们在G2和G3的unseen tool的embedding分析，Logic-based 对齐前后的与retriever见过的retriever data的Kl 散度对比

\subsection{Noise Analysis (Q4)}

To analyze the robustness of \texttt{LoSemB} to noisy training data, we evaluate the retrieval performance under scenarios where 10\% of instruction-tool pairs in the training set are incorrect, simulating real-world scenarios where tool annotations may be imperfect. We compare the performance of \texttt{LoSemB} against baseline methods on ToolBench (I2) under both clean and noisy training conditions.

\begin{figure}[h]
\vspace{-2mm}
  \centering
  \includegraphics[width=0.98\linewidth]{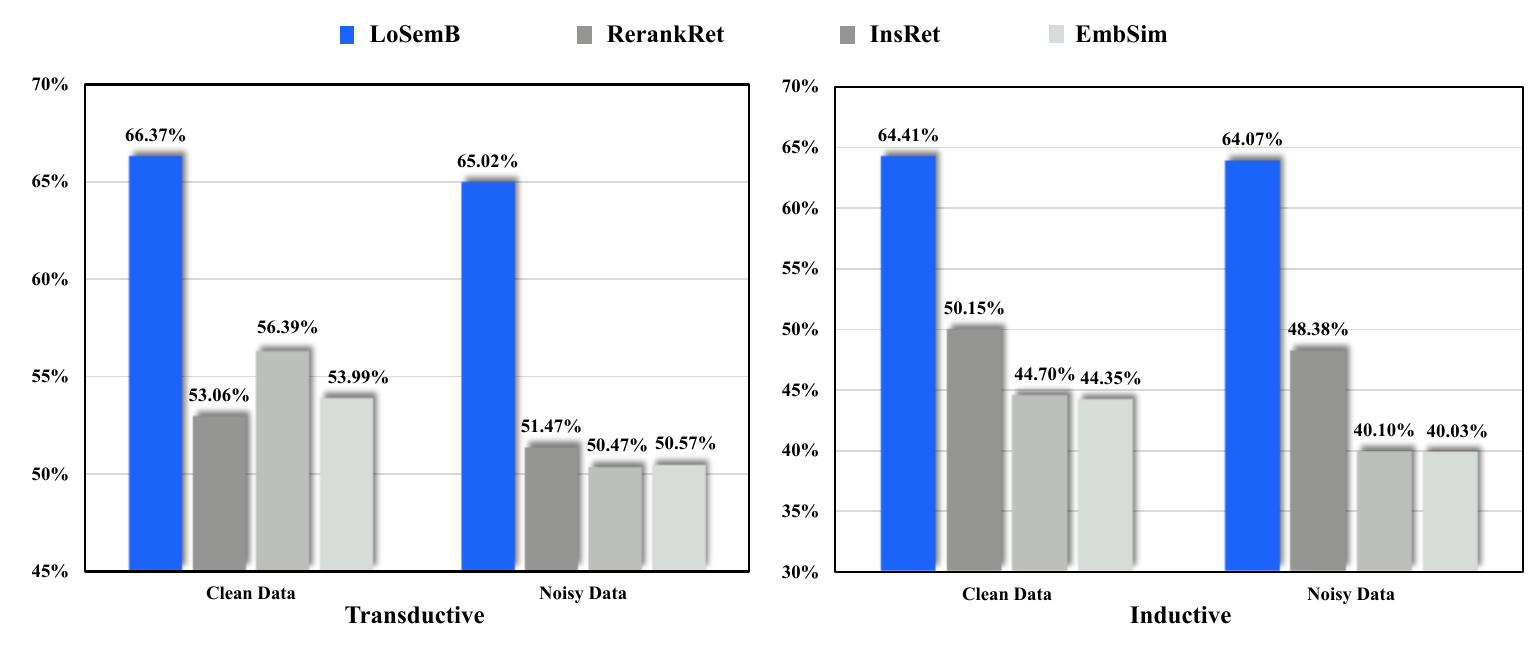}
  \vspace{-4mm}
\caption{Retrieval results (\%) of different baselines under transductive and inductive settings. Clean data indicates no label noise.}
\vspace{-2mm}
\label{fig:noise}
\end{figure}

\textbf{Obs. 7. \texttt{LoSemB} demonstrates superior robustness to label noise compared to baseline methods.} As shown in Figure \ref{fig:noise}, \texttt{LoSemB} exhibits minimal performance drops of only 1.35\% and 0.34\% under 10\% label noise, while InsRet suffers absolute drops of 5.92\% and 4.60\%. This robustness is attributed to the dual verification mechanism in Relational Augmented Retrieval.

\section{Conclusion}

The main objective of this paper is to tackle the tool retrieval task in the inductive setting. Although retrieval-based methods have shown excellent performance, they still face two challenges: the large distribution shift and the sensitivity of semantic-only retrieval when handling unseen tools. To this end, we propose a logic-based tool retrieval framework, named \texttt{LoSemB}, which integrates latent logical information into the retrieval process. We first employ a logic-based embedding alignment module to mitigate the large distribution shift of unseen tools, and then implement a relational augmented retrieval mechanism that integrates logical pruning and graph-enhanced similarity matching. Extensive experiments in both transductive and inductive settings demonstrate the strong performance of \texttt{LoSemB}. Looking ahead, \texttt{LoSemB} introduces a new paradigm for tool retrieval, paving the way for advanced techniques like sophisticated GNNs and complex graph structures to further enhance tool learning in evolving web tool ecosystems.

\section{Acknowledgments}

The work described in this paper was fully supported by a grant from the Innovation and Technology Commission of the Hong Kong Special Administrative Region, China (Project No. ITS/263/24FP).

\bibliographystyle{ACM-Reference-Format}
\balance
\bibliography{main}

\appendix 
\section{Related Work}

\subsection{Tool Retrieval} 

Recently, LLMs have shown excellent abilities in many tasks. Meanwhile, it becomes increasingly vital to equip LLMs with external tools~\cite{du2024anytool,mu2024query}. In this regard, tool retrieval plays a crucial role, with two main research directions emerging: 

\begin{table*}[t]\small
	\centering
	\vspace{-2mm}
	\caption{Dataset statistics on ToolBench, UltraTool, and APIGen datasets across varying proportions of unseen tools. Note that the test set remains consistent across all ratios.}
	\label{tab:setting}
	\resizebox{0.95\textwidth}{!}{
	\begin{tabular}{cccccccccccc} 
	\toprule
	\multirow{2}{*}{\textbf{Type}} & \multirow{2}{*}{\textbf{Ratio}}  
	 &\multicolumn{2}{c}{\textbf{ToolBench (I1)}} 
	 &\multicolumn{2}{c}{\textbf{ToolBench (I2)}} 
	 &\multicolumn{2}{c}{\textbf{ToolBench (I3)}} 
	 &\multicolumn{2}{c}{\textbf{UltraTool}} 
	 &\multicolumn{2}{c}{\textbf{APIGen}} \\ 
	\cmidrule(lr){3-4} \cmidrule(lr){5-6} \cmidrule(lr){7-8} \cmidrule(lr){9-10} \cmidrule(lr){11-12}
	& & Instruction & Tool & Instruction & Tool & Instruction & Tool & Instruction & Tool & Instruction & Tool \\ 
	\midrule[0.6pt]
	\multirow{4}{*}{Train}
	&0\% &86,643 &10,439 &84,207 &13,142 &25,044 &1,605 &5,682 &2,032 &59,800&3,579\\ 
	&10\% &84,355 &10,330 &82,104 &13,070 &19,095 &1,573 &4,163 &2,020 &55,272&3,541\\
	&20\% &82,136 &10,221 &79,100 &12,998 &14,783 &1,541 &3,153 &2,009 &51,241&3,502\\ 
	&30\% &80,616 &10,112 &76,795 &12,926 &11,261 &1,509 &2,490 &1,998 &47,403&3,464\\ 
	\midrule
	\multirow{1}{*}{Test}
	&all &792 &1,089 &568 &720 &216 &317 &149 &112 &200&386\\ 
	\bottomrule
	\end{tabular}
	}
	\vspace{-1mm}
\end{table*}
\textbf{Token-based methods.} Token-based methods~\cite{hao2023toolkengpt,wang2024toolgen} represent each tool as a specific token and generate the relevant tools. For example, ToolGen~\cite{wang2024toolgen} implements a comprehensive four-stage process: tool virtualization, tool memorization, retrieval training, and end-to-end agent-tuning. This approach involves a resource-intensive training process. When tools are integrated as tokens, additional tokens must be added and their corresponding documentation fine-tuned into the model's parameters.

\textbf{Retrieval-based methods.} Retrieval-based methods~\cite{zhuang2025linearrag,zhang2025survey,xiao2025reliablereasoningpathdistilling,zhang2025faithfulrag} search relevant tools from the large tool repositories. These methods include both sparse (\textit{e.g.}, BM25 \cite{robertson2009probabilistic}) and dense retrievers, mainly using PLMs. A number of studies have built upon this baseline method~\cite{hong2024knowledge,xiang2025use,xiao2025laglogicaugmentedgenerationcartesian}. For example, ToolRetriever~\cite{qintoolllm} uses instruction-tool pairs to train PLMs on domain-specific data. QTA \cite{zhang2024data} addresses low-resource scenarios by using direct preference optimization (DPO) to train LLMs. However, these transductive approaches face challenges from distribution shifts and the vulnerability of similarity-based retrieval in inductive settings. Although Re-invoke~\cite{chen2024re} proposes a zero-shot tool retrieval framework, its performance is worse than supervised approaches.

% \subsection{Inductive Learning}

% Recent inductive learning studies have critically evaluated the inductive reasoning capabilities of LLMs. While Chen et al.~\cite{bowen2024comprehensive} identifies emergent problem-solving skills in LLMs, it also highlights their inconsistency under distribution shifts. Cheng et al.~\cite{cheng2024inductive} further critiques the conflation of inductive and deductive reasoning in LLMs, revealing vulnerabilities in handling unseen logical patterns. To address this, Rytting et al.~\cite{rytting2021leveraging} leverages LLMs’ inherent inductive bias for textual reasoning. Though relevant, these efforts focus on textual or symbolic reasoning rather than tool retrieval, where logical dependencies between tools must guide generalization. Graph-based inductive representation learning inspires knowledge graph reasoning frameworks like~\cite{galkintowards}. These existing tool retrieval methods often assume static repositories or rely on simple similarity metrics. Our work bridges this gap by unifying logical reasoning with retrieval mechanisms, enabling robust generalization to evolving tool ecosystems without costly retraining.

\subsection{Graph Convolutional Networks}

In recent years, several convolutional neural network architectures for learning over graphs have been proposed~\cite{zhou2025each,zhang2024knowgpt}, which have been shown to effectively model relationship features, relational reasoning~\cite{shengyuan2023differentiable,hong2024next,nsea,chen2025dontneedprebuiltgraphs}, and combinatorial generalization~\cite{luo2024fairgt,luo2025fairgp,gllongsurvey2025,chen2024llm4ea}. For instance, TransConv~\cite{lai2019transconv} further advances this paradigm by incorporating textual interactions between user pairs in social networks. UniMP~\cite{shi2021masked} conceptually unifies feature propagation and label propagation through a unified message passing framework. These GNN methods demonstrate effectiveness in semi-supervised classification tasks. Therefore, we adopt GNNs to extract logical features. In this work, we build upon LightGCN~\cite{he2020lightgcn}, which is particularly suitable for capturing logical information between instructions and tools. Specifically, our approach combines logical features and text features to achieve embedding alignment for unseen tools through feature propagation and aggregation. 

\section{Implementation Details}
\label{app:exp_details}

\subsection{Data Statistics}
\label{app:data}

To accurately assess the impact of unseen tools on tool retrieval performance, we conducted experiments under both transductive and inductive settings. In the transductive setting, we filtered the test set to remove instructions involving tools not present in the training data. For the inductive setting, we randomly selected 10\%, 20\%, and 30\% of test tools as unseen tools and excluded them from the training data. The detailed statistics for both settings across all datasets are presented in Table \ref{tab:setting}.

\subsection{Hyperparameters}

We set the training epochs for the retriever to 5 with a learning rate of $2 \times 10^{-5}$ and warmup steps of 500. We carefully tune the layer number $L$ among $\{2, 3, 4\}$, parameter $J$ among $\{3, 4, 5, 6, 7\}$, and parameter $I$ among $\{2, 3, 4, 5\}$. For the GCN component, we set the training epochs to 200 with a learning rate of $1 \times 10^{-3}$ and a weight decay of $1 \times 10^{-4}$.

\section{Additional Experiments}
\label{app:add_exp}

\subsection{Efficiency analysis (Q5)}

To address Q5, we compare the efficiency of \texttt{LoSemB} against training-based baselines on ToolBench (I2) under the transductive setting where 30\% of tools are unseen. Our efficiency analysis spans three stages: (1) the indexing stage; (2) the incremental indexing (Incr. Indexing) stage for processing and incorporating unseen tools; and (3) the inference stage for retrieval. We observe the following:

\begin{table}[htbp]
\centering
\caption{Efficiency and performance comparison of training-based baselines on ToolBench (I2) with 30\% unseen tools. Notably, Perf. denotes the average retrieval performance across R@3, R@7, P@3, and P@7.}
\label{tab:efficiency}
\resizebox{\columnwidth}{!}{%
\begin{tabular}{lcccc}
\toprule
\textbf{Method} & \textbf{Indexing (h)} & \textbf{Incr. Indexing (min)} & \textbf{Inference (ms)} & \textbf{Perf. (\%)} \\
\midrule
EmbSim & 8.42 & 668.33 & 11.37 & 53.99 \\
InsRet& 14.53 & 930.33 & 21.14 & 56.39 \\
RerankRet& 24.86 & 2148.80 & 4373.82 & 53.77\\
\texttt{LoSemB} & 8.58 & 5.66 & 12.38 & 64.41 \\
\bottomrule
\end{tabular}%
}
\vspace{-3mm}
\end{table}

\textbf{Obs. 8. LoSemB achieves superior efficiency and effectiveness in handling unseen tools.} As shown in Table~\ref{tab:efficiency}, LoSemB exhibits comparable initial indexing time (8.58h vs. 8.42h) and inference time (12.38ms vs. 11.37ms) to the traditional EmbSim approach. However, the key distinction emerges when processing unseen tools through incremental indexing. EmbSim requires computationally expensive retriever retraining, consuming 668.33 minutes (over 11 hours). In contrast, LoSemB leverages its logic-guided semantic bridging approach to process unseen tools in merely 5.66 minutes—a 118$\times$ speedup. This dramatic efficiency gain, coupled with a substantial performance improvement (64.41\% vs. 53.99\%), makes LoSemB the most practical choice for deployments demanding strong performance in frequently updated tool repositories, where scalability and cost control are critical.

\subsection{Impact of Sentence Embeddings (Q6)}

In this section, we study the effectiveness of different sentence embeddings on retrieval performance in \texttt{LoSemB}. We compare Ada Embedding, BERT-base~\citep{devlin2019bert} (110M parameters), RoBERTa-base~\citep{liu2019roberta} (125M parameters), GTR-T5-base~\citep{ni2022large} (110M parameters), and BERT-large-uncased~\citep{devlin2019bert} (336M parameters) on the ToolBench (I2) dataset. Notably, except for Ada Embedding, all other models are retriever backbones fine-tuned on instruction-tool pairs. The retrieval performance is evaluated under both transductive and inductive (30\% unseen tools) settings.

\begin{figure}[h]
    \centering
    \begin{subfigure}[b]{0.44\linewidth}
        \centering
        \includegraphics[width=\textwidth]{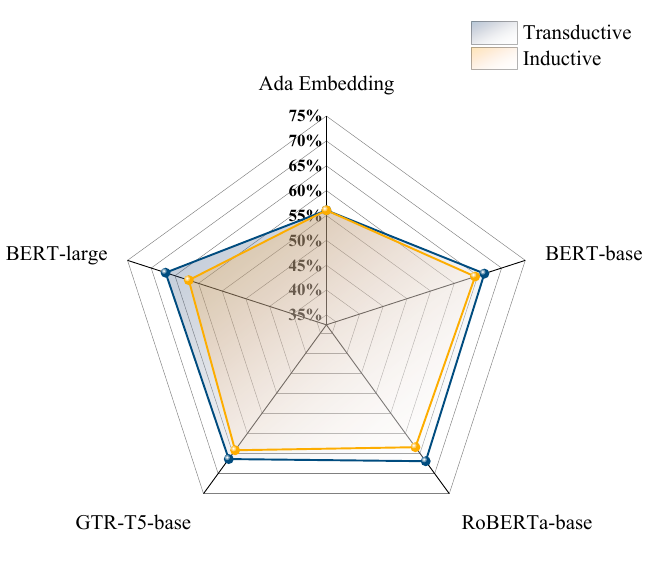}
        \caption{Impact of Different Sentence Embeddings}
        \label{fig:backbone}
    \end{subfigure}
    \hfill
    \begin{subfigure}[b]{0.55\linewidth}
        \centering
        \includegraphics[width=\textwidth]{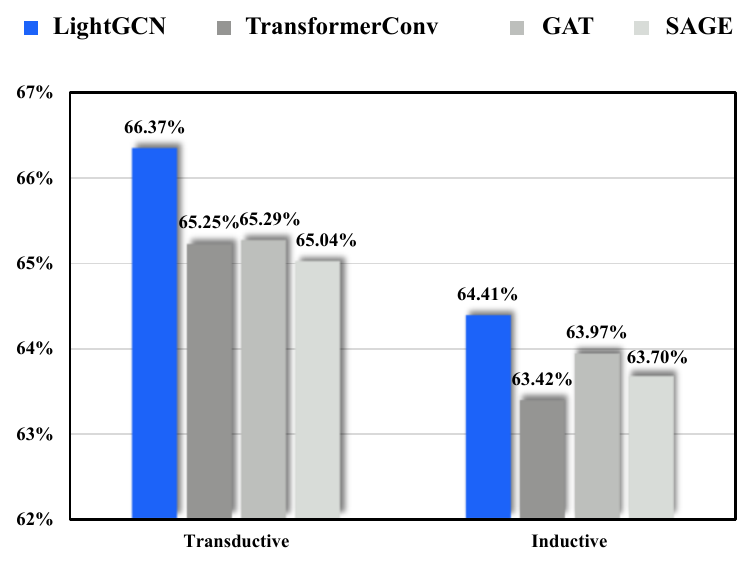}
        \caption{Impact of Different GNNs}
        \label{fig:gnn}
    \end{subfigure}
    \caption{\textbf{(a)} Impact of different sentence embeddings, \textbf{(b)} Impact of GNNs. All results are evaluated on ToolBench (I2).}
    \label{fig:gnn_backbone}
    \vspace{-3mm}
\end{figure}

\textbf{Obs. 9. BERT-base demonstrates the best trade-off between performance and robustness to unseen tools among all evaluated sentence embeddings.} Based on the findings presented in Table~\ref{fig:gnn_backbone} (a), we observe that while all sentence embedding models achieve relatively comparable performance, BERT-base stands out by demonstrating both strong effectiveness in the transductive setting and superior robustness in the inductive setting with unseen tools. Notably, even Ada Embedding without domain-specific training achieves competitive results, demonstrating the effectiveness of our framework. Consequently, we adopt BERT-base as the default embedding model in all experiments.

\subsection{Impact of Different GNNs (Q7)}
In this section, we investigate the impact of different GNN architectures on the retrieval performance of \texttt{LoSemB}. We compare four representative GNN models: GraphSAGE~\citep{hamilton2017inductive}, GAT~\citep{velivckovic2018graph}, TransformerConv~\citep{dwivedi2020generalization}, and LightGCN~\citep{he2020lightgcn} on the ToolBench (I2) and ToolBench (I3) datasets. The evaluation is conducted under both transductive and inductive (30\% unseen tools) settings.

\textbf{Obs. 10. LightGCN consistently achieves the best performance across different settings among all evaluated GNN architectures.} Based on the findings presented in Figure~\ref{fig:gnn_backbone} (b), we observe that while all GNN architectures demonstrate competitive performance, LightGCN exhibits superior results in the majority of experimental configurations. However, LightGCN's simplified design, which removes feature transformation and nonlinear activation while focusing on neighborhood aggregation, proves particularly effective for capturing tool-instruction relationships in our logic-guided semantic space. Consequently, we adopt LightGCN as the default GNN architecture in all experiments.

\subsection{Parameter Sensitivity (Q8)}

We investigate the impact of key hyperparameters in \texttt{LoSemB}. Experiments are evaluated with the average scores of R@3, R@7, P@3, and P@7 in transductive and inductive settings on ToolBench (I2).

\begin{figure}[tbp]
  \centering
  \begin{subfigure}[b]{0.48\textwidth}
    \includegraphics[width=\textwidth]{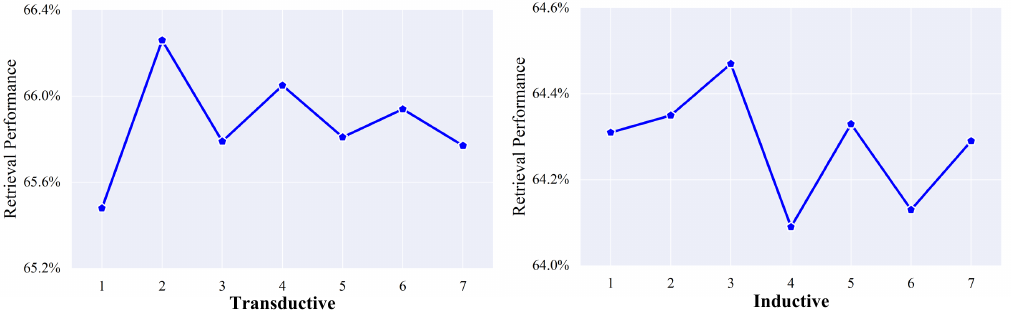}
    \caption{Impact of Layer Depth $L$}
    \label{fig:sub1}
  \end{subfigure}
  \hfill
  \begin{subfigure}[b]{0.48\textwidth}
    \includegraphics[width=\textwidth]{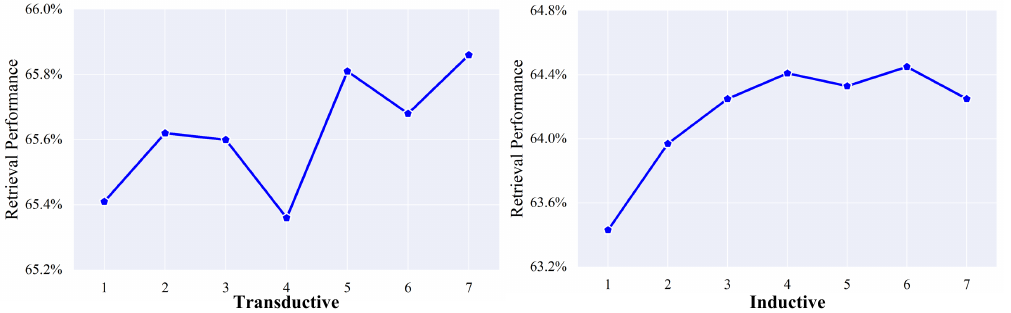}
    \caption{Impact of $J$}
    \label{fig:sub2}
  \end{subfigure}
  \hfill
  \begin{subfigure}[b]{0.48\textwidth}
    \includegraphics[width=\textwidth]{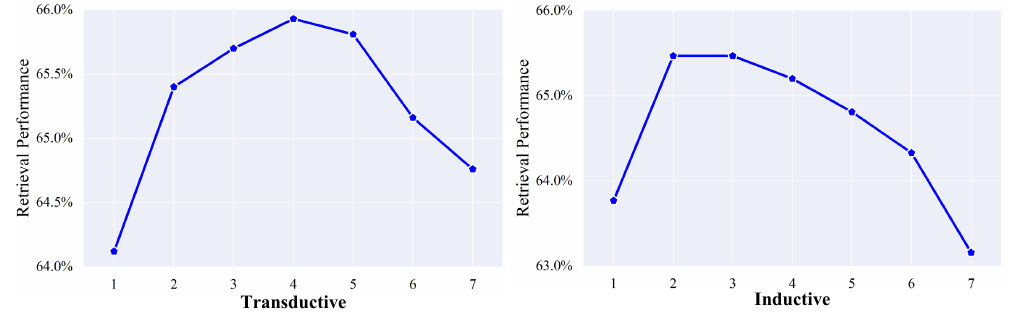}
    \caption{Impact of $I$}
    \label{fig:sub2}
  \end{subfigure}
  \caption{Parameter analysis of \texttt{LoSemB}. (a) shows the dependency of \texttt{LoSemB} performance on layer depth $L$. (b) illustrates the influence of $J$. (c) examines the effect of $I$.}
  \label{fig: para}
  \vspace{-5mm}
\end{figure}

\textbf{Obs. 11. Moderate layer depth yields optimal performance, while excessive depth leads to over-smoothing and degraded retrieval quality.} Based on the findings presented in Figure~\ref{fig: para}~(a), we observe that incorporating logical information to generate graph embeddings yields superior performance compared to relying solely on semantic embeddings. This demonstrates the effectiveness of our logic-guided semantic bridging approach. However, deeper layers enable richer neighborhood information aggregation, excessively large $L$ values can be harmful to the effectiveness of \texttt{LoSemB} due to the over-smoothing problem. Consequently, we adopt a moderate layer depth in all experiments to balance the benefits of multi-hop reasoning with the preservation of feature distinctiveness.

\textbf{Obs. 12. Obtaining logical information from a moderate number of similar instructions effectively improves performance, while excessive nodes introduce irrelevant information and degrade results.} Based on the findings presented in Figure~\ref{fig: para}~(b), we observe that obtaining logical information from similar instructions and tools is effective, with performance improving as the number increases. This demonstrates the effectiveness of our logical feature transfer mechanism. However,  more nodes are added, the improvement slows down and eventually drops, as less relevant nodes bring in unhelpful information that interferes with the model's judgment. Consequently, we adopt a moderate $J$ value to fully leverage logical information from similar instructions while avoiding noise introduction.

\textbf{Obs. 13. The logical tool set size must balance recall and precision, as oversized sets introduce functionally mismatched tools.} Based on the findings presented in Figure~\ref{fig: para}~(c), we observe that as $I$ increases, the logical tool set $\mathcal{T}_\mathrm{logic}$ selected based on the most similar instructions grows larger, initially improving performance. This indicates that expanding the candidate set helps increase the likelihood of recalling relevant tools. However, once $I$ becomes too large, performance begins to decline as larger sets tend to include tools that are textually similar but functionally different, introducing errors that negatively affect performance. Consequently, we adopt an appropriate $I$ value to achieve the optimal balance between expanding the candidate scope and maintaining functional relevance.

\subsection{Case Study on Retrieved Tools (Q9)} 

\begin{table}[h]
	\centering
	\caption{Selected example from ToolBench showing top-1 tool retrieval results.}
	\label{tab:case}
	\small
	\begin{tabular}{p{0.95\columnwidth}}
		\toprule
		\textbf{Instruction:} I'm planning a family gathering and I want to make some non-alcoholic cocktails for the kids. Can you suggest some fun and colorful mocktail recipes? Also, provide me with the contact details of a brewery that offers non-alcoholic beer options. \\
		\midrule
		\textbf{EmbSim:} [Food, Tasty, recipes/detail] \\
		\midrule
		\textbf{LoSemB:} [Food, The Cocktail DB, Detailed Cocktail Recipe by ID] \\
        \midrule
        \textbf{Label:} [Food, The Cocktail DB, Detailed Cocktail Recipe by ID] \\
		\bottomrule
	\end{tabular}
\vspace{-3mm}
\end{table}

In this section, we demonstrate how tool retrieval benefits from \texttt{LoSemB} using the ToolBench dataset as examples. Our analysis focuses on two key aspects: the logic-based embedding alignment module and the relational augmented retrieval mechanism.

\textbf{Obs. 14. The synergy between logic-based embedding alignment and relational augmented retrieval enables \texttt{LoSemB} to overcome the limitations of semantic-only methods.} Based on the case presented in Table~\ref{tab:case}, we observe that EmbSim retrieves a generic recipe tool based on textual similarity to ``recipes,'' failing to capture the specific requirement for cocktails. In contrast, \texttt{LoSemB} correctly identifies ``The Cocktail DB'' tool through two complementary mechanisms working in tandem. First, the logic-based embedding alignment module learns the semantic relationships among instructions involving terms like ``mocktail,'' ``cocktail,'' and ``recipes'' through the instruction-tool graph structure, enabling the model to understand that mocktail instructions require cocktail-specific tools rather than general recipe databases. Second, the relational augmented retrieval mechanism identifies similar instructions within the cocktail domain (e.g., ``I want to surprise my partner with a special cocktail. Can you suggest a unique and romantic cocktail recipe?'') and leverages their associated tools to narrow down the search space to cocktail-relevant tools. This demonstrates the practical value of our approach in real-world tool recommendation scenarios where domain-specific understanding is crucial.

% while transformer-based embeddings achieve comparable performance in the transductive setting, their robustness to unseen tools varies significantly. BERT-base maintains superior performance in the inductive setting, outperforming even larger models. This suggests that moderate model capacity, when combined with our logic-guided semantic bridging approach, provides better generalization to unseen tools. Consequently, we adopt BERT-base as the default embedding model in all experiments due to its optimal balance of computational efficiency and robustness.
% % 在一个两个数据集不同backbone的效果

\end{document}